\def\paperID{0000}
\def\confName{ICCV}
\def\confYear{2025}

\def\authorBlock{
Jiaqi Wu$^{1,2}$\footnotemark[1]\hspace{1em}, 
Yaosen Chen$^{1,2}$\footnotemark[2]\hspace{1em}, 
Shuyuan Zhu$^{1}$\footnotemark[2] \\
$^{1}$University of Electronic Science and Technology of China, Chengdu, China \\
$^{2}$Sobey Media Intelligence Laboratory, Chengdu, China \\

{\tt\small wjq4dpm@std.uestc.edu.cn, chenyaosen@sobey.com, eezsy@uestc.edu.cn} \\
}

\newif\ifreview 
\newif\ifarxiv \newcommand{\arxiv}{\arxivtrue}
\newif\ifcamera 
\newif\ifrebuttal

\arxiv 

\pdfoutput=1
\documentclass[10pt,twocolumn,letterpaper]{article}
\ifreview \usepackage[review]{cvpr} \fi
\ifarxiv \usepackage[pagenumbers]{cvpr} \fi
\ifrebuttal \usepackage[rebuttal]{cvpr} \fi
\ifcamera \usepackage{cvpr} \fi

\usepackage{graphicx}	
\usepackage{amsmath}	
\usepackage{amssymb}	
\usepackage{booktabs}
\usepackage{times}
\usepackage{microtype}
\usepackage{epsfig}
\usepackage{caption}
\usepackage{float}
\usepackage{placeins}
\usepackage{color, colortbl}
\usepackage{stfloats}
\usepackage{enumitem}
\usepackage{tabularx}
\usepackage{xstring}
\usepackage{multirow}
\usepackage{xspace}
\usepackage{url}
\usepackage{subcaption}
\usepackage{xcolor}
\usepackage[hang,flushmargin]{footmisc}

\ifcamera \usepackage[accsupp]{axessibility} \fi

% ...

\ifarxiv  \fi

\newcommand{\R}[1]{{%
    \textbf{%
        \ifstrequal{#1}{1}{\textcolor{red}{R#1}}{%
        \ifstrequal{#1}{2}{\textcolor{blue}{R#1}}{%
        \ifstrequal{#1}{3}{\textcolor{magenta}{R#1}}{%
        \ifstrequal{#1}{4}{\textcolor{teal}{R#1}}{%
                           \textcolor{cyan}{R#1}%
        }}}}%
    }%
}}

\usepackage{xr-hyper}
\usepackage{setspace}
\makeatletter
\newcommand*{\addFileDependency}[1]{
  \typeout{(#1)}
  \@addtofilelist{#1}
  \IfFileExists{#1}{}{\typeout{No file #1.}}
}

\makeatother
\newcommand*{\myexternaldocument}[1]{
    \externaldocument{#1}
    \addFileDependency{#1.tex}
    \addFileDependency{#1.aux}
}

\definecolor{cvprblue}{rgb}{0.21,0.49,0.74}
\usepackage[pagebackref,breaklinks,colorlinks,allcolors=cvprblue]{hyperref}
\usepackage[capitalize]{cleveref}
\crefname{section}{Sec.}{Secs.}
\crefname{table}{Table}{Tables}
\crefname{figure}{Fig.}{Figs.}

\ifarxiv \crefname{appendix}{App.}{Apps.}
\else \crefname{appendix}{Suppl.}{Suppls.} \fi

\usepackage{adjustbox}
\unless\ifarxiv \myexternaldocument{_supplementary} \fi

\begin{document}
\title{GeoMVD: Geometry-Enhanced Multi-View Generation Model Based on Geometric Information Extraction}
\author{\authorBlock}
\maketitle

\renewcommand{\thefootnote}{\fnsymbol{footnote}}
\footnotetext[2]{Corresponding Authors.}
\footnotetext[1]{Work done during an internship at SobeyMIL.}

\begin{abstract}

Multi-view image generation holds significant application value in computer vision, 
particularly in domains like 3D reconstruction, virtual reality, and augmented reality. 
Most existing methods, which rely on extending single images, face notable computational 
challenges in maintaining cross-view consistency and generating high-resolution outputs. 
To address these issues, we propose the Geometry-guided Multi-View Diffusion Model, 
which incorporates mechanisms for extracting multi-view geometric information and 
adjusting the intensity of geometric features to generate images that are both 
consistent across views and rich in detail. Specifically, we design a multi-view 
geometry information extraction module that leverages depth maps, normal maps, 
and foreground segmentation masks to construct a shared geometric structure, ensuring 
shape and structural consistency across different views. To enhance consistency and 
detail restoration during generation, we develop a decoupled geometry-enhanced attention 
mechanism that strengthens feature focus on key geometric details, thereby improving 
overall image quality and detail preservation. Furthermore, we apply an adaptive 
learning strategy that fine-tunes the model to better capture spatial relationships 
and visual coherence between the generated views, ensuring realistic results. Our model 
also incorporates an iterative refinement process that progressively improves the output 
quality through multiple stages of image generation. Finally, a dynamic geometry 
information intensity adjustment mechanism is proposed to adaptively regulate the 
influence of geometric data, optimizing overall quality while ensuring the naturalness 
of generated images. More details can be found on the project 
page: \url{https://sobeymil.github.io/GeoMVD.com/}.
\end{abstract}

\section{Introduction}
\label{sec:intro}

\begin{figure*}[ht]
    \centering
    \includegraphics[width=0.85\textwidth]{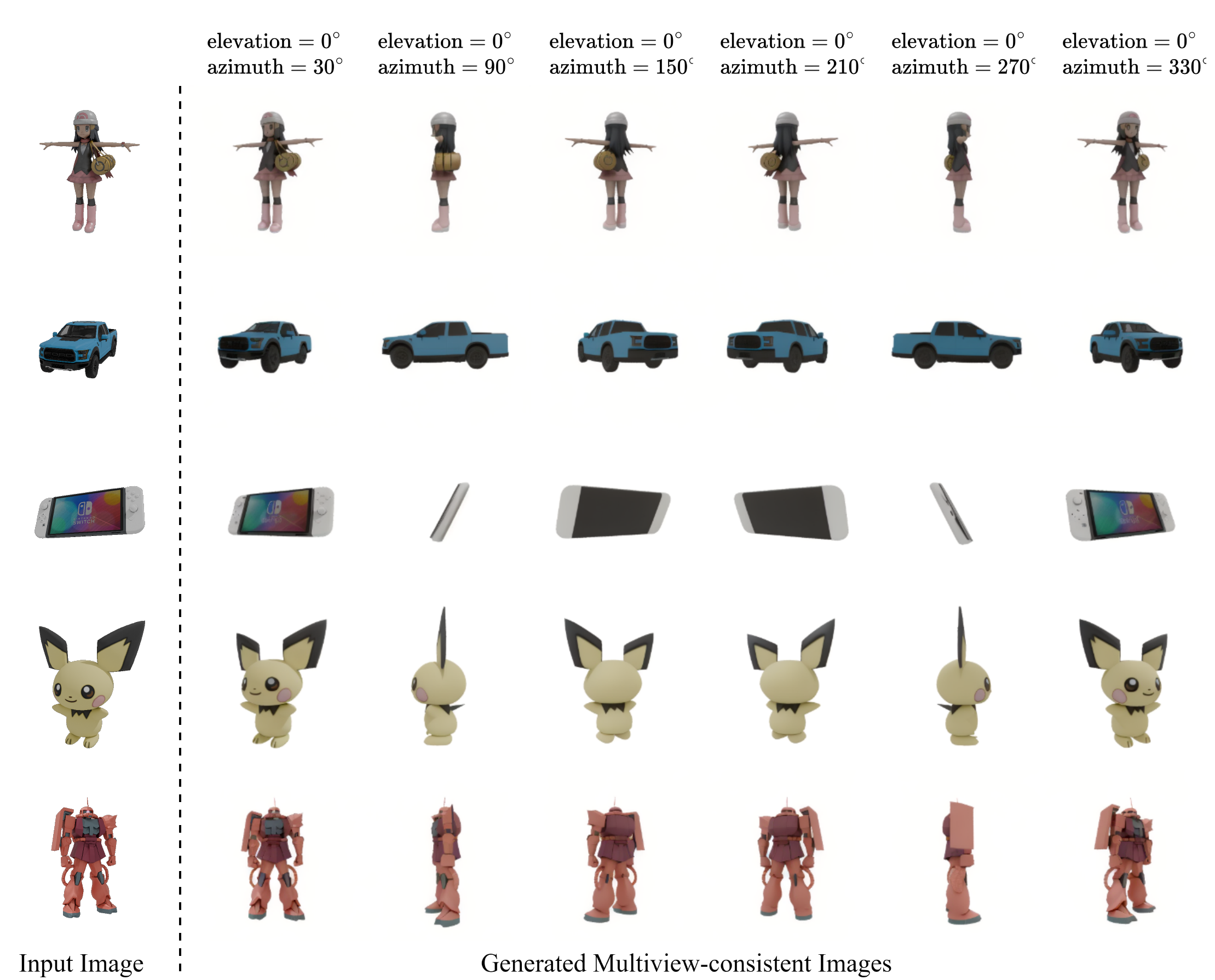}  
    \caption{GeoMVD performs excellently on single-image–driven multi-view three-dimensional generation, is compatible with real photographs, synthesized images, and two-dimensional illustrations, and maintains cross-view consistency and high quality.}
    \label{fig:top}
\end{figure*}

Multi-view image generation is a fundamental task in the field of computer vision, playing 
a crucial role in various applications such as 3D modeling, virtual reality, augmented 
reality, robot perception, and simulation. With the rapid development of "text-to-image" 
diffusion models, significant progress has been made in single-view image generation 
\cite{rombach2022ldm,nichol2021glide,saharia2022imagen,ramesh2022dalle2, betker2023dalle3,
podell2023sdxl,esser2024sd3,batifol2025flux}. 
These models are capable of generating clear, detail-rich images from textual descriptions, 
greatly expanding the boundaries of automated content creation. However, extending this 
technology to multi-view image generation and integrating text, images, and 3D data within 
a unified framework still faces numerous challenges. Multi-view image generation holds 
great promise in providing strong technical support for 3D reconstruction, virtual reality, 
and autonomous driving. However, ensuring cross-view consistency, geometric accuracy, and 
overcoming the computational challenges posed by high-resolution images and large-scale 3D 
datasets remain critical issues in current research.

Existing multi-view diffusion models generally achieve image generation tasks by fine-tuning 
"text-to-image" models 
\cite{shi2023mvdream,wang2023imagedream,shi2023zero123++,deng2023mvdiff,tang2024mvdiffusion++,
liu2023syncdreamer,huang2024epidiff,gao2024cat3d,long2024wonder3d,li2024era3d,kant2024spad,
zheng2024free3d}. 
These methods model 3D consistency between images by applying attention across multiple views. 
However, as the resolution of images and the scale of 3D data increase, the computational cost 
becomes significantly higher. In particular, as model sizes grow and resolution requirements 
escalate, training and inference processes become computationally expensive, making it difficult 
to achieve the desired image quality. Moreover, despite some advanced methods being trained on 
large-scale datasets, the lack of high-quality 3D training data increases the difficulty of 
optimization, leading to gaps in the generated multi-view image quality in both detail and 
consistency. Therefore, maintaining image consistency, geometric accuracy, and detail richness 
while handling large-scale data is a critical issue in current multi-view image generation research.

To address these bottlenecks, we propose a geometry-guided multi-view diffusion model. The core 
idea is to first estimate depth and normal maps from a single input image and combine them with 
foreground segmentation to construct a continuous, edge-preserving proxy surface via bilateral 
normal integration. Then, we render the multi-view geometry images from the target viewpoints 
based on this proxy surface, aligning them naturally within a shared geometric framework, thus 
preventing boundary misalignment, occlusion inconsistencies, and detail drift. The overall pipeline 
uses "input image, optional text, and multi-view geometry images" as conditions to extract 
appearance details, cross-modal semantics, and geometric structures, which are jointly utilized 
at each diffusion step to ensure semantic consistency, geometric accuracy, and appearance fidelity 
in the generated images.

In terms of feature fusion, we propose a decoupled geometry-enhanced attention mechanism: in addition 
to self-attention, a parallel geometry attention branch is constructed, with denoising features as the 
query. The image and geometry conditions are used separately to calculate attention, and then adaptively 
fused with learnable weights. This significantly strengthens the shape and occlusion constraints across 
views without sacrificing texture restoration. Considering that geometry is estimated from a single image 
and may contain errors, we introduce two robustness mechanisms: first, a geometry feature mask based on 
the cosine similarity between the target viewpoint and the input viewpoint, which adaptively reduces the 
constraint strength in unreliable views; and second, dynamic adjustment of geometry attention weights 
according to diffusion steps. Early steps have stronger weights to correct noisy outputs, while later 
steps reduce the weights to avoid excessive constraints that may damage details. During training, we only 
add low-rank adaptive fine-tuning parameters to the attention layers, achieving high-resolution generation 
with minimal trainable parameters and enhanced efficiency and stability.

In summary, our contributions are as follows:
\begin{enumerate}
    \item We propose a geometry-guided multi-view diffusion framework and multi-view 
    geometry information extraction module: Depth and normal maps are estimated from 
    a single image and combined with segmentation to construct a proxy surface, which 
    is rendered into multi-view geometry images, providing a shared geometric constraint 
    that systematically enhances cross-view consistency and detail stability.
    
    \item We design a triple-condition collaborative denoising process for cross-modal 
    semantics, geometric structure, and image appearance: These three conditions are 
    jointly constrained via cross-attention and subsequent modules, ensuring semantic 
    consistency, structural coherence, and appearance fidelity at each diffusion step.
    
    \item We propose a decoupled geometry-enhanced attention mechanism: This mechanism 
    employs a parallel dual-branch attention structure, with one branch processing image 
    features and the other handling geometric features. The final output is obtained by 
    weighted fusion of the attention results from both branches, ensuring consistency and 
    detail preservation across multiple views without over-relying on either geometry or 
    image details.

    \item We introduce a robust inference strategy: This strategy uses a geometry feature 
    mask based on viewpoint differences and dynamically adjusts geometry attention weights 
    according to diffusion steps. This improves stability and naturalness, ensuring 
    high-quality multi-view generation with minimal computational cost.

\end{enumerate}
\section{Related Work}
\label{sec:related}

\paragraph{Text-to-Image Diffusion Models.} Text-to-image generation has made 
continuous breakthroughs in the evolution of diffusion models, advancing from basic 
alignment and efficiency optimization to high resolution, strong controllability, 
and cross-scene adaptability \cite{ho2020ddpm,song2020ddim,dhariwal2021guideddiffusion,
ho2022cfg,rombach2022ldm,ramesh2022dalle2,podell2023sdxl,esser2024sd3,batifol2025flux,
cao2025hunyuanimage,zhang2025tvg}. Guided diffusion \cite{dhariwal2021guideddiffusion} and 
classifier-free guidance\cite{ho2022cfg} established the foundation for text-conditioned 
image generation; DALL-E 2 \cite{ramesh2022dalle2} leveraged CLIP \cite{radford2021clip} 
to achieve precise text-image alignment; latent diffusion models\cite{rombach2022ldm} 
enhanced efficiency through latent space diffusion; and Stable Diffusion XL \cite{podell2023sdxl} 
optimized high-frequency details with a two-stage cascaded architecture. Recent 
developments have introduced the Flow Matching\cite{lipman2022flow} technique as an 
efficient alternative framework, enabling smooth noise-to-image transformation through 
continuous dynamic flow mapping, significantly reducing sampling steps and improving 
generation stability. Stable Diffusion 3(SD3)\cite{esser2024sd3} and FLUX are both based on the Diffusion Transformer\cite{peebles2023dit} 
architecture and utilize flow matching techniques to improve text-image alignment accuracy. 
SD3 integrates two CLIP encoders and one T5\cite{raffel2020t5} encoder, optimizing text condition 
processing and image detail generation, thereby enhancing prompt adherence and image generation 
precision. In contrast, FLUX\cite{batifol2025flux} enhances image style consistency, multi-round 
editing task responsiveness, and multimodal context processing through its "Kontext" model series, 
significantly improving the stability and editability of high-resolution generation.

\paragraph{Multiview Generation Based on T2I Models.} Multiview generation methods
\cite{shi2023mvdream,deng2023mvdiff,tang2024mvdiffusion++,huang2024epidiff,gao2024cat3d,
liu2023syncdreamer,long2024wonder3d,li2024era3d,kant2024spad,zheng2024free3d}
have expanded text-to-image models by leveraging large-scale 3D datasets \cite{yu2023mvimgnet,
deitke2023objaverse,lin2025objaverse++}
 For example, MVDream \cite{shi2023mvdream} integrates camera embedding and extends 
 self-attention from 2D to 3D to enable cross-view coherence; SPAD \cite{kant2024spad} 
 enhances cross-view attention by applying epipolar constraints. Zero123 \cite{liu2023zero123} 
 presents a conditional diffusion method to generate consistent multiviews from a single 
 view, improving cross-view coherence though still limited under high resolution and 
 complex scenes; Zero123++ \cite{shi2023zero123++} refines this by introducing more efficient 
 feature extraction strategies to enhance precision and style consistency in multiview 
 generation. Era3D \cite{li2024era3d} introduces an epipolar-aligned row-level self-attention 
 mechanism, improving stability in multiview generation; SyncDreamer \cite{liu2023syncdreamer} 
 proposes a synchronized multiview diffusion model that leverages 3D-aware attention for 
 improved geometric consistency and image quality. NVS-Adapter \cite{jeong2024nvs} introduced 
 a plug-and-play module for pretrained T2I models, specifically for novel view synthesis 
 from a single image. It uses view-consistency cross-attention to align local features 
 across views and global semantic conditioning to align the semantic structure of generated 
 views with a reference view, achieving geometrically consistent multiview synthesis without 
 full model fine-tuning. Building on this, MV-Adapter \cite{huang2025mv_adapter} further 
 extends the concept, offering a plug-and-play adapter that enables pretrained T2I models 
 to perform multiview generation efficiently by preserving their feature space and structure. 
 MV-Adapter significantly enhances the generation performance and application diversity by 
 ensuring better compatibility with various multiview tasks.

Despite these advances, many methods still require large-scale parameter updates to pretrained 
T2I models, limiting compatibility with model derivatives. This study introduces GeoMVD, a 
geometry-guided multiview diffusion model. GeoMVD introduces a multiview geometry extraction 
module, a triplet-condition collaborative denoising process, a decoupled geometry-enhanced 
attention mechanism, and a geometry-information intensity adjustment mechanism. These 
innovations significantly improve image quality, consistency, and detail fidelity, 
outperforming existing methods in cross-view consistency, geometric precision, and 
generation quality.
\section{Method}
\label{sec:method}

\begin{figure*}[ht]
    \centering
    \includegraphics[width=0.98\textwidth]{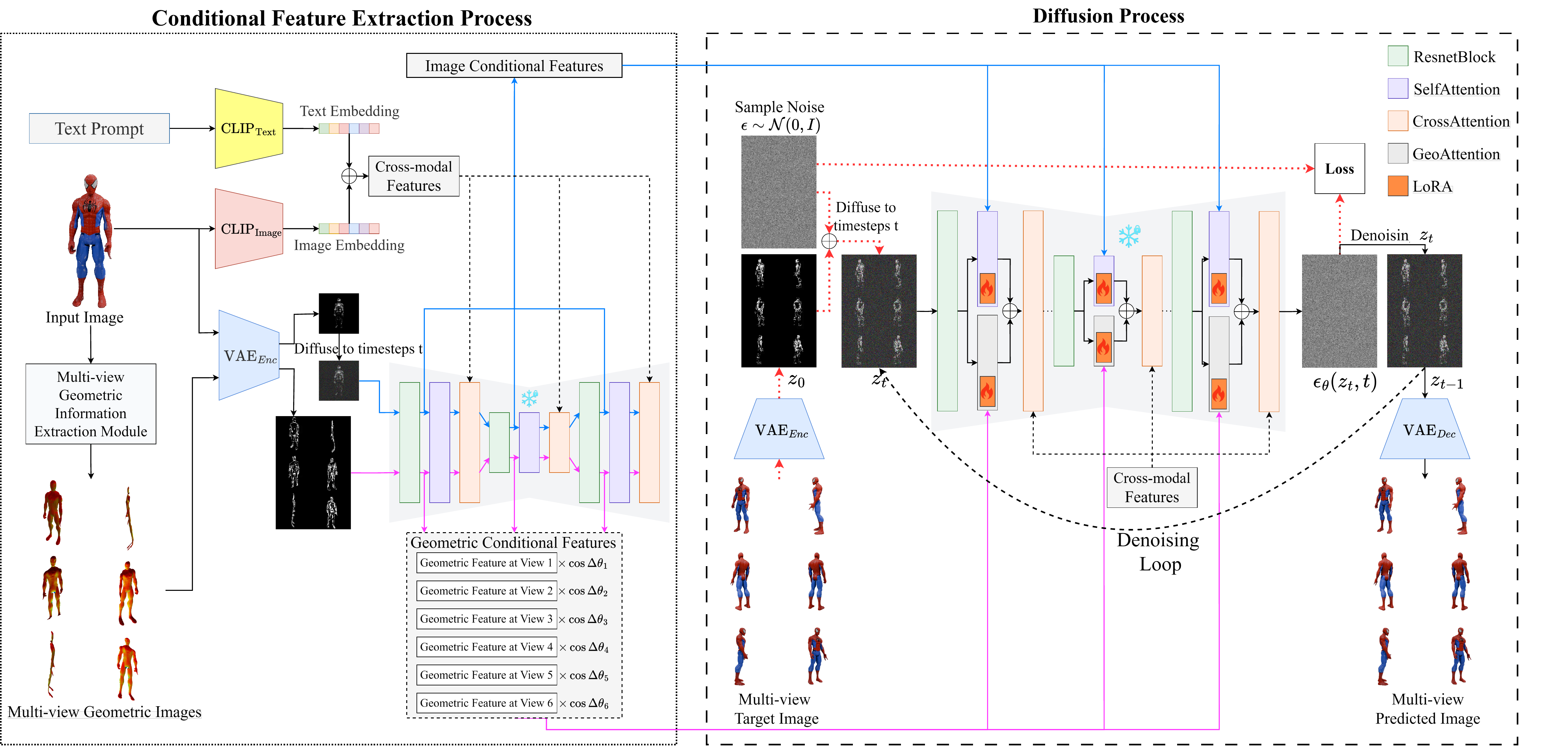}  
    \caption{The process of the multi-view diffusion model guided by geometric information.}
    \label{fig:pipe}
\end{figure*}

The overall process of the multi-view diffusion model proposed in this study is illustrated in 
Fig.\ref{fig:pipe}. The input to the diffusion model consists of: an image containing the target 
object, an optional textual description, and multi-view geometric images. These geometric images 
are generated from the multi-view geometric information extraction module (Section 3.1), which 
predicts geometric features from a single image under multiple views. During the conditional 
feature extraction process (Section 3.2), these three inputs are transformed into cross-modal, 
multi-view geometric, and image conditional features using pre-trained models such as VAE, CLIP, 
and DiffusionUnet. The cross-modal conditional features are input to the cross-attention layer of 
the U-Net, while image and multi-view geometric features are processed by the decoupled 
geometry-enhanced attention module (Section 3.3). Additionally, the geometric information 
intensity modulation mechanism (Section 3.4) dynamically adjusts the strength of the geometric 
guidance during the generation process to ensure stable and effective constraints.

\subsection{Multi-view Geometric Information Extraction Module}

Traditional multi-view diffusion models generate multi-view images directly from a single image. 
However, due to the absence of reliable 3D geometric priors, these models often struggle with 
structural consistency across views, leading to issues such as boundary mismatches, occlusion 
inconsistencies, and detail drift.

\begin{figure}[ht]
    \centering
    \includegraphics[width=0.48\textwidth]{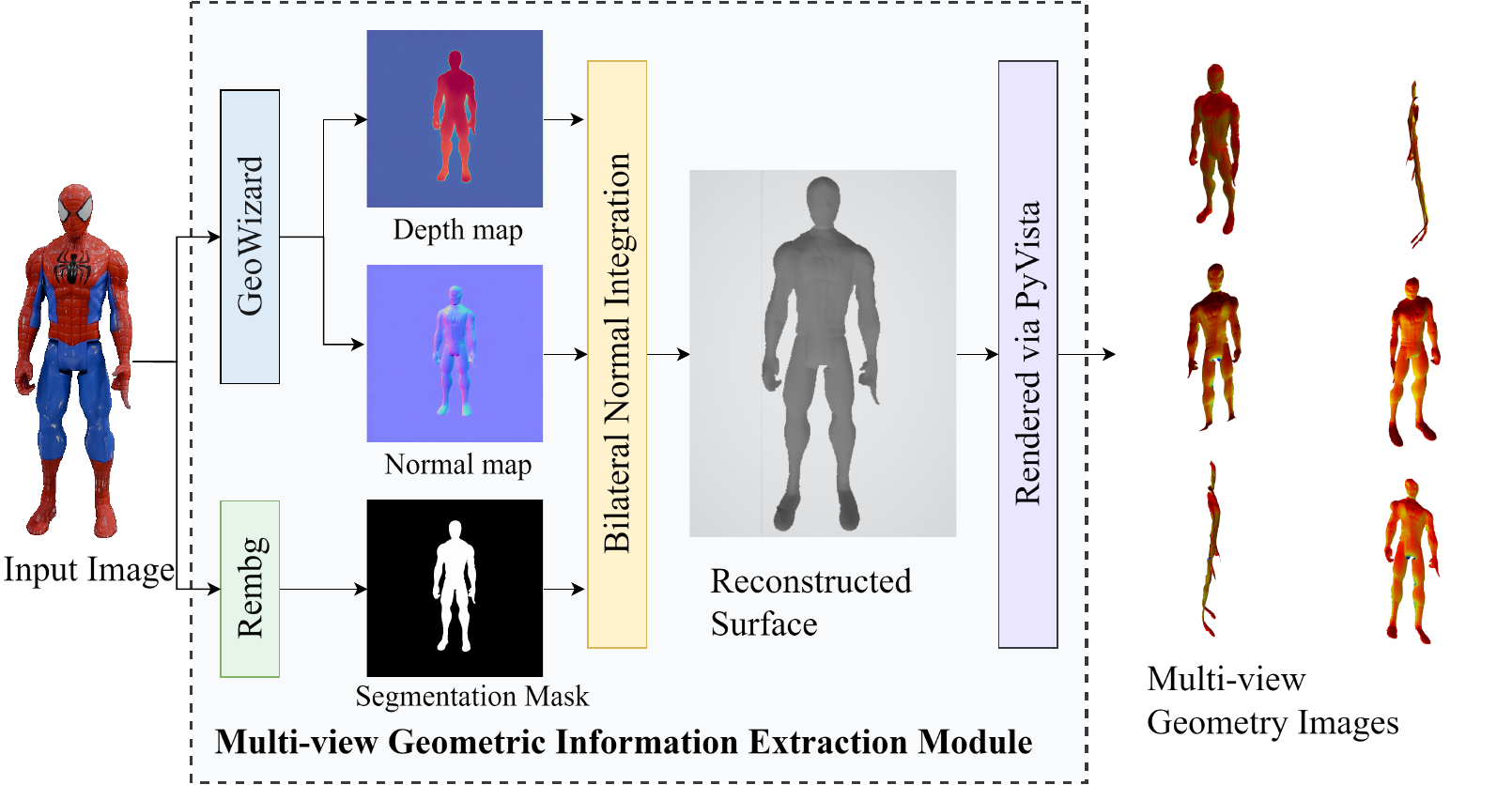}  
    \caption{Schematic Diagram of the Multi-view Geometric Information Extraction Module.}
    \label{fig:mv_info}
\end{figure}

To address these challenges, we propose the Multi-view Geometric Information Extraction Module, 
which shifts the generation task from the image domain to view-independent proxy surfaces. By 
establishing a unified geometric representation of the object’s shape and topology, we ensure 
all views share the same geometric structure. As shown in Fig.\ref{fig:mv_info}, we use 
GeoWizard \cite{fu2024geowizard} to estimate depth and normal maps from the input image, 
and Rembg \cite{qin2020u2} to obtain the foreground segmentation mask. The Bilateral Normal 
Integration (BiNI) algorithm \cite{cao2022bini} then fuses these inputs into a continuous, 
edge-preserving surface (BiNI Proxy Surface, BPS). PyVista \cite{pyvista_website} is then 
used to render the BPS from the target viewpoint and color it using a normalized normal-based 
color mapping, producing multi-view geometric images. These images serve as geometric conditional 
features to guide the multi-view diffusion model's generation process.

This approach ensures that all views are constrained within a unified geometric framework, 
maintaining the consistency of the object's overall shape, the relative positions of its parts, 
and the occlusion relationships. Furthermore, normal-based detail cues (such as edges, contours, 
and subtle curvature variations) provide persistent high-frequency information for the denoising 
process, effectively reducing detail drift and hallucinated textures. Ultimately, our method 
ensures cross-view consistency while preserving rich and stable details.

\subsection{Conditional Feature Extraction Process}

Generating high-quality multi-view results from a single image requires satisfying three key 
constraints: semantic consistency (consistent object, attributes, and style across views), 
geometric consistency (coherent shape, structure, and occlusion relationships across views), 
and appearance fidelity (color and texture consistent with the input image). To meet these 
constraints, we propose a tri-conditional collaborative 
constraint method that jointly utilizes cross-modal, geometric, and image information, ensuring 
both consistency and detail preservation at each denoising step.

As illustrated in Fig.\ref{fig:pipe}, our method extracts three types of conditional features:

\begin{itemize}

\item \textbf{Cross-modal Conditional Features}: The text prompt and the input image are separately 
input into the text encoder and image encoder of CLIP, generating two embedding vectors. These 
vectors are then linearly fused to form the cross-modal feature. This feature is input into the 
cross-attention layer of the U-Net, ensuring consistency in semantic categories, attributes, and 
style across different views.

\item \textbf{Image Conditional Features}: The input image is encoded into a latent representation 
using a VAE. Noise is added to this latent representation based on the noise level of the current 
iffusion step. The noisy latent representation is then passed into the U-Net, and input 
representations are extracted from all self-attention layers to form the image conditional features. 
The role of the image conditional features is to provide the shape, detail, and texture information 
of the object, helping the model recover the true appearance of the object and ensuring consistency 
in structure, shape, and details with the input image.

\item \textbf{Geometric Conditional Features}: Multi-view geometric images are encoded into latent 
representations via a VAE without adding noise. These features are directly input into the U-Net, 
and input representations are extracted from all self-attention layers to obtain the geometric 
conditional features. The role of the geometric conditional features is to ensure that the object 
maintains consistent shape, structure, and occlusion relationships across different views, providing 
stable geometric constraints and preserving the 3D feel and spatial coherence of the object.

\end{itemize}

These three types of conditional features are processed through cross-attention and decoupled 
geometry-enhanced attention layers. The cross-modal conditional features stabilize semantic and 
style consistency across different views. The geometric features provide consistency constraints 
across views, suppressing perspective distortions and structural mismatches. The image features 
are responsible for recovering the appearance and mid-to-high frequency details. To ensure that 
the conditions and the generated target share the same noise scale at each diffusion step, noise 
is added to the image latent representation, providing matching denoising gradient signals that 
help stabilize the recovery of color and texture. On the other hand, the geometric features carry 
high-frequency and strong constraint shape information, and adding noise would weaken their 
stability and constraint strength across steps. Therefore, they are kept noise-free to consistently 
apply clear and stable geometric constraints during the denoising process. Through the collaborative 
effect of these three conditions, the model ensures semantic consistency, structural coherence, 
and appearance fidelity in the multi-view generation process.

\subsection{Decoupled Geometry-Enhanced Attention Mechanism}

\begin{figure}
    \centering
    \includegraphics[width=0.46\textwidth]{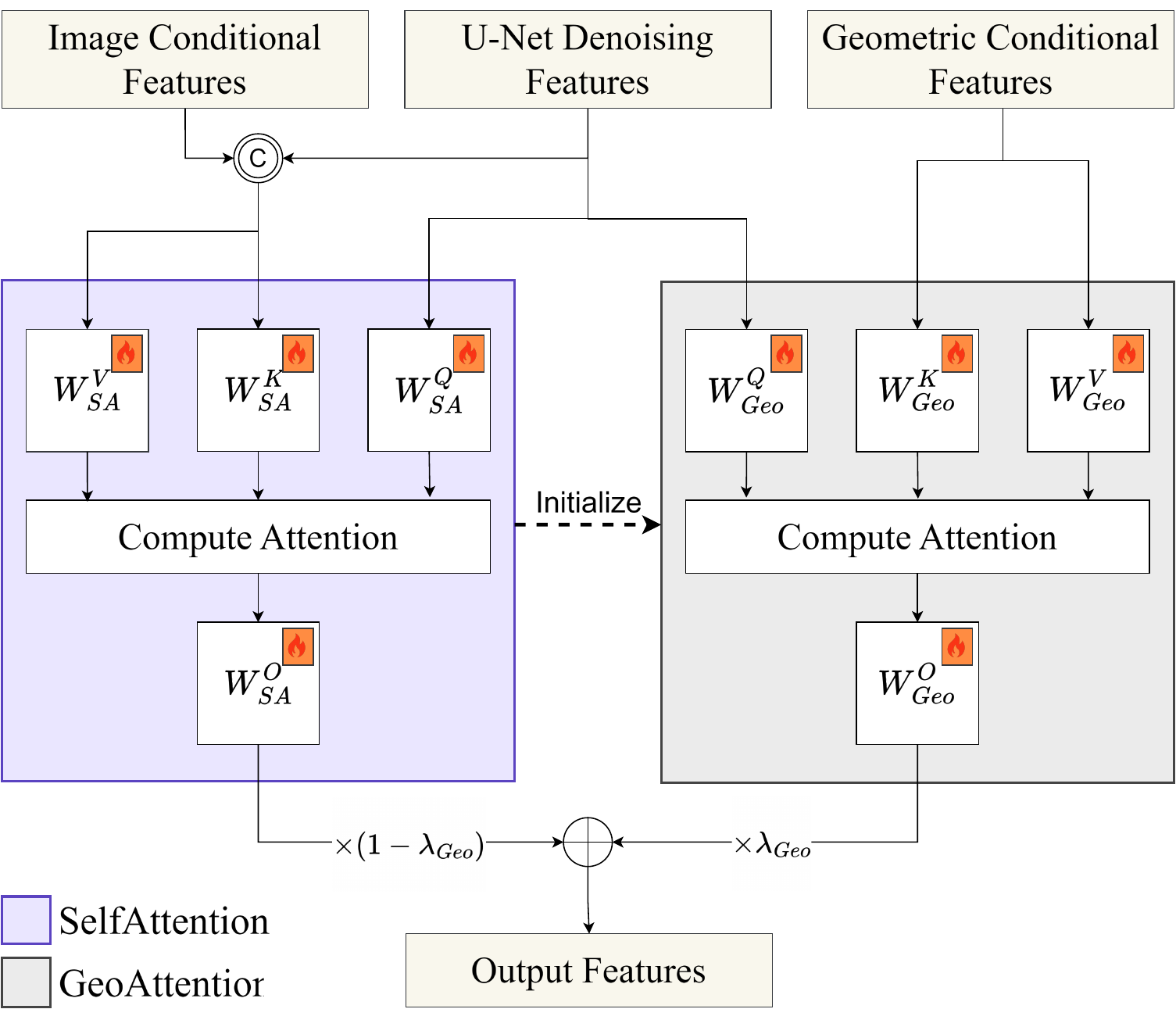}  
    \caption{Decoupled Geometry-Enhanced Attention Mechanism}
    \label{fig:atten}
\end{figure}

As shown in the left block diagram of Fig.\ref{fig:atten}, the decoupled geometry-enhanced 
attention module improves upon the self-attention mechanism of the diffusion model U-Net. 
This module introduces a parallel attention structure by initializing the geometric attention 
layers using the same weights as the self-attention layers, allowing parallel processing of 
both image features and geometric features.

Specifically, the module takes three inputs: image conditional features $F_{\text{img}}$, 
U-Net denoising features $F_{\text{unet}}$, and geometric conditional features 
$F_{\text{Geo}}$. In the model, U-Net denoising features $F_{\text{unet}}$ serve as the 
query (Q) input for both attention modules. The image conditional features $F_{\text{img}}$ 
and denoising features $F_{\text{unet}}$ are concatenated along the length dimension to form 
the key (K) and value (V) inputs for the self-attention layer, while geometric conditional 
features $F_{\text{Geo}}$ are used as the key (K) and value (V) inputs for the geometric 
attention layer.

In this structure, the two attention modules independently compute attention scores for 
image and geometric features. The final feature output $O$ is then obtained by weighted 
fusion, ensuring consistency and detail recovery for the generated image across multiple 
views.

The detailed computation process is as follows:

\begin{enumerate}
    \item \textbf{Self-Attention Module (Self Attention):} The image conditional features $F_{\text{img}}$ and 
    U-Net denoising features $F_{\text{unet}}$ are concatenated and linearly transformed using $W_{\text{SA}}$ to 
    compute the query (Q), key (K), and value (V). The attention output is then calculated through the self-attention mechanism:
    
    \begin{align}
        Q_{\text{SA}} &= F_{\text{unet}} W_{\text{SA}}^{Q}, \\
        K_{\text{SA}} &= (F_{\text{img}} \oplus F_{\text{unet}}) W_{\text{SA}}^{K}, \\
        V_{\text{SA}} &= (F_{\text{img}} \oplus F_{\text{unet}}) W_{\text{SA}}^{V}, \\
        A_{\text{SA}} &= \text{Softmax}\left( \frac{Q_{\text{SA}} K_{\text{SA}}^T}{\sqrt{d_k}} \right) V_{\text{SA}} 
    \end{align}

    \item \textbf{Geometric Attention Module (Geo Attention):} For the geometric conditional features, the query (Q) 
    is computed using U-Net denoising features $F_{\text{unet}}$, while the key (K) and value (V) are derived from 
    the geometric conditional features $F_{\text{Geo}}$. The corresponding linear transformations use the geometric 
    attention module's weight matrix $W_{\text{GA}}$:
    \begin{align}
        Q_{\text{GA}} &= F_{\text{unet}} W_{\text{SA}}^{Q}, \\
        K_{\text{GA}} &= F_{\text{Geo}} W_{\text{GA}}^{K}, \\
        V_{\text{GA}} &= F_{\text{Geo}} W_{\text{GA}}^{V}, \\
        A_{\text{GA}} &= \text{Softmax}\left( \frac{Q_{\text{GA}} K_{\text{GA}}^T}{\sqrt{d_k}} \right) V_{\text{GA}} 
    \end{align}

    \item \textbf{Weighted Fusion of Outputs:} Finally, the attention outputs of the image conditional and geometric 
    conditional features are fused by weighted summation to produce the final feature output $O$. The weight coefficient 
    $\lambda_{\text{geo}}$ is used to balance the contributions of both features:
    \begin{equation}
        O = (1 - \lambda_{\text{geo}}) \cdot A_{\text{SA}} W_{\text{SA}}^{O} + \lambda_{\text{geo}} \cdot A_{\text{GA}} W_{\text{GA}}^{O} 
    \end{equation}
\end{enumerate}

This architecture efficiently processes both image and geometric features simultaneously, 
ensuring consistency across multiple views and maintaining object shape and detail fidelity. 
By balancing the weighted fusion, the model avoids over-reliance on either geometric constraints 
or image details, resulting in more refined and realistic multi-view image generation.

\subsection{Geometric Information Intensity Modulation Mechanism}

\begin{figure}
    \centering
    \includegraphics[width=0.45\textwidth]{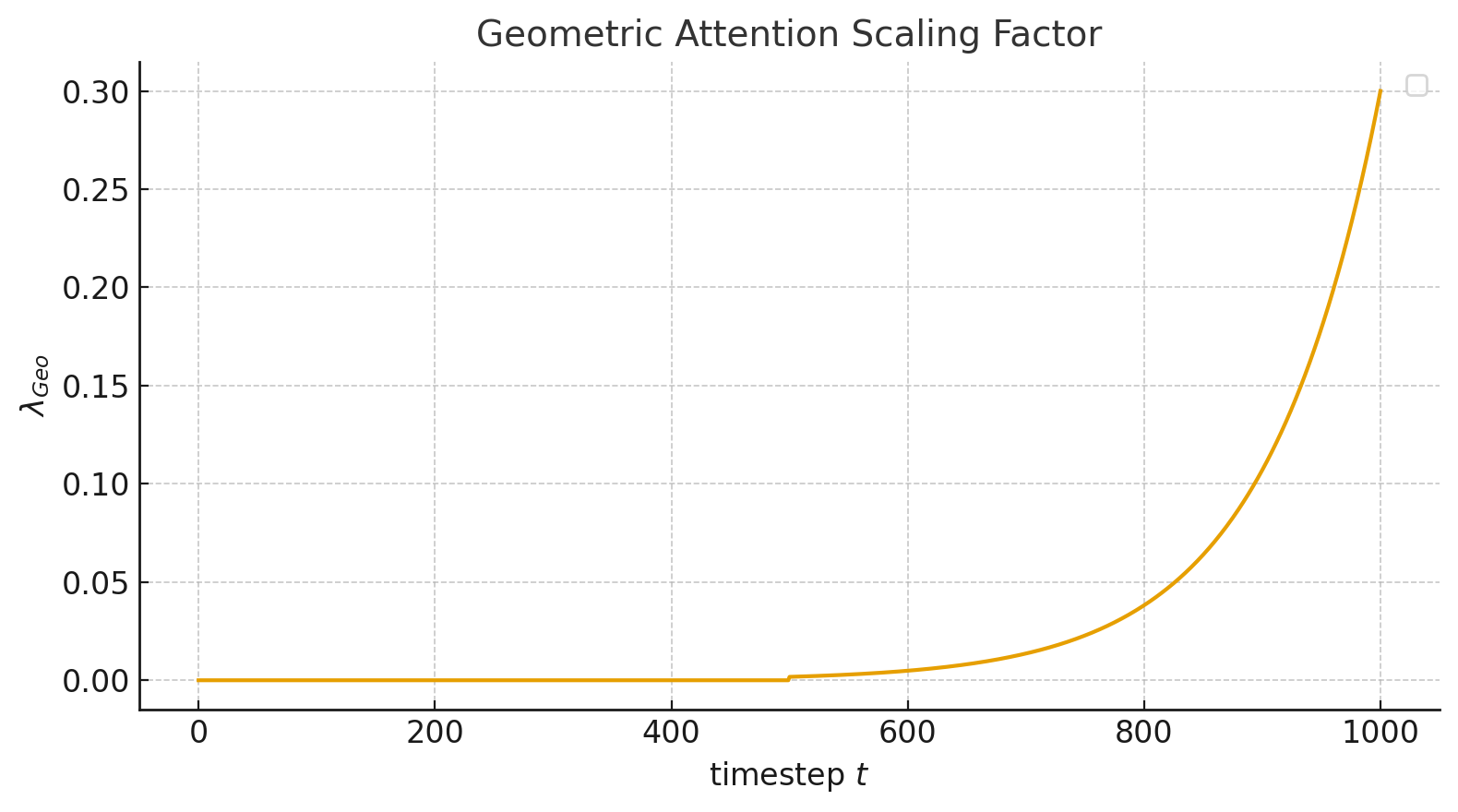}  
    \caption{Variation of Geometric Attention Scaling Factor with Diffusion Steps}
    \label{fig:scale_factor}
\end{figure}

Based on observations during the experimental process, we applied two adjustments to the 
geometric conditional features: the Multi-view Geometric Feature Mask and Dynamic Geometric 
Attention Strength Scheduling.

\paragraph{Multi-view Geometric Feature Mask:} Since geometric information is entirely 
computed from a single image, some views may contain erroneous information. To address 
this issue, we introduce a cosine similarity-based mechanism that dynamically adjusts 
the intensity of multi-view geometric features based on the difference between the 
target and input views, using it as a mask for correction. Specifically, the further 
a view deviates from the main viewpoint, the less accurate the information becomes, 
requiring the adjustment of the strength for each view. The adjustment is given by:
   \begin{equation}
   \tilde{f}_{\text{Geo}}^{n} = \cos(\Delta \theta_n) \cdot f_{\text{Geo}}^n
   \end{equation}
   where $f_{\text{Geo}}^n$ represents the geometric feature for the $n$-th view, 
   and $\Delta \theta_n$ is the deviation between the $n$-th view and the main viewpoint. 
   The cosine of the angle, $\cos(\Delta \theta_n)$, is used to scale the feature 
   intensity, resulting in the adjusted geometric feature $\tilde{f}_{\text{Geo}}^n$.

   Through this adjustment mechanism, we can dynamically scale the geometric feature 
   intensity based on the deviation of each view from the main viewpoint, effectively 
   avoiding the interference of erroneous information and ensuring the consistency and 
   accuracy of the generated geometric features across different views.

\paragraph{Dynamic Geometric Attention Strength Scheduling:} During the experimental process, 
we observed that in the early stages of the denoising loop, a larger geometric attention scaling 
factor $\lambda_{\text{geo}}$ effectively guides the early denoising process due to the high noise 
in the generated image. However, as the denoising progresses, the geometric branch may interfere 
with the generated result due to modality differences. Therefore, the geometric attention scaling 
factor $\lambda_{\text{geo}}$ should gradually decrease with the timestep. The decay process is 
shown in Fig.\ref*{fig:scale_factor}, where the factor decays from 0.3 to $10^{-5}$ following a 
geometric series (i.e., geometric progression), and the scaling factor is set to 0 for the first 500 steps.

    \begin{equation}
    \lambda_{\text{geo}}(t) = 
    \begin{cases}
    0, & \text{if } t \in [0, T//2) \\
    1e^{-5} \cdot \left(\frac{0.3}{1e^{-5}}\right)^{\frac{t - T//2}{T//2}}, & \text{if } t \in [T//2, T]
    \end{cases}
    \end{equation}

where $T$ is the total number of denoising steps, and $t$ is the current denoising step.

Through these two adjustments, this study effectively controls the influence of 
geometric information, thereby improving the quality of multi-view generated images. 
These improvements ensure that at each stage of the generation process, geometric 
information can precisely guide the denoising process without excessively affecting 
the details of the final generated image.

\section{Experiment}
\label{sec:related}

\textbf{Dataset} 
We trained GeoMVD on a subset of the Objaverse\cite{deitke2023objaverse} dataset. To construct the training images, 
we rendered one single-view input image and six ground-truth images using the Cycles\cite{blender_cycles} 
engine in Blender. The azimuth and elevation angles of the single-view input image were 
set to 10°, while the ground-truth images were rendered with azimuth angles evenly 
distributed from 30° to 330°, with a fixed elevation angle of 0°. All rendered images 
had a resolution of $320 \times 320$ and were in PNG format with an alpha channel.

To evaluate the performance of GeoMVD, we conducted experiments on the Google Scanned 
Object\cite{downs2022GSO} (GSO) dataset, which is a widely recognized benchmark in the 3D generation field. 
This dataset is commonly used for performance comparison and validation of related 
algorithms. Additionally, we tested our method on out-of-domain images collected from 
the internet to demonstrate its generalization ability. Similarly to previous methods 
, we removed the background from these out-of-domain images 
using Rembg and centered the objects.

\textbf{Implementation Details} 
Our implementation is based on the open-source multi-view diffusion model Zero123++\cite{shi2023zero123++}, 
with optimizations built upon this foundation. During training, we integrated LoRA\cite{hu2022lora} into 
the self-attention and geometric attention layers of the U-Net, rather than training 
the entire U-Net, which significantly reduced the number of trainable parameters. 
The final model had approximately 2.8M trainable parameters. This strategy effectively 
improved both training efficiency and model performance. We trained GeoMVD on four GeForce 
RTX 3090 GPUs with a batch size of 8 and performed 
10,000 steps of training. The AdamW optimizer was used, along with a cosine annealing 
learning rate scheduler. The learning rate started at $5 \times 10^{-5}$ and gradually 
decayed to $1.25 \times 10^{-5}$, with restarts every 1,000 steps. The entire training 
process took approximately 15 hours.

\subsection{Experimental Results}

We compare GeoMVD with Era3D, Zero123++, and MV-Adapter (based on SDXL) in generating multi-view 
images. All methods generate images from six fixed viewpoints, but the viewpoints differ across 
the methods. Specifically, GeoMVD, Zero123++, and MV-Adapter use horizontal angles of 
\(\{30^{\circ}, 90^{\circ}, 150^{\circ}, 210^{\circ}, 270^{\circ}, 330^{\circ}\}\), while Era3D 
generates images from horizontal angles of \(\{0^{\circ}, 45^{\circ}, 90^{\circ}, 180^{\circ}, 270^{\circ}, 315^{\circ}\}\). 
The elevation angles for GeoMVD, Era3D, and MV-Adapter are all \(0^{\circ}\), while Zero123++ generates 
images with varying elevation angles of \(\{30^{\circ}, -20^{\circ}, 30^{\circ}, -20^{\circ}, 30^{\circ}, -20^{\circ}\}\). 
Furthermore, GeoMVD and Era3D use white backgrounds, while Zero123++ and MV-Adapter utilize gray backgrounds.

\subsubsection{Qualitative Comparison}

\begin{figure*}
    \centering
    \includegraphics[width=0.97\textwidth]{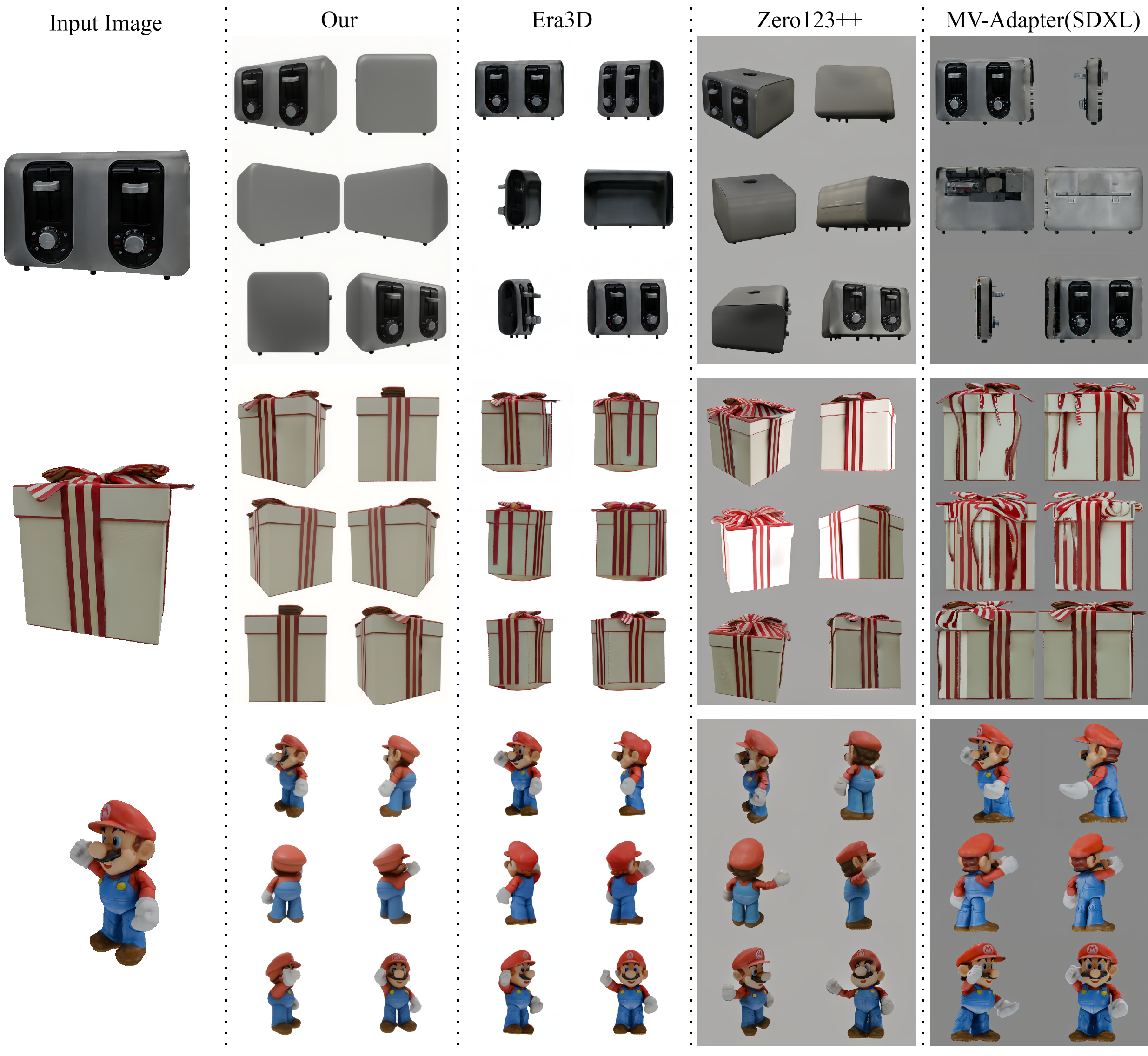}  
    \caption{Comparison of Generation Results on the GSO Dataset between GeoMVD and Other Methods.}
    \label{fig:cmp_GSO}
\end{figure*}

Fig.\ref{fig:cmp_GSO} presents the generated results of GeoMVD, Era3D, Zero123++, and MV-Adapter on 
the GSO dataset. The comparison clearly shows that GeoMVD outperforms the other methods in generating 
consistent multi-view images from a single input. For instance, when generating images of the toaster, 
gift box, and Mario models, our algorithm produces clear and consistent multi-view images, maintaining 
superior detail and consistency compared to Era3D and Zero123++, which exhibit artifacts or detail loss 
in some views. MV-Adapter (based on SDXL) generates blurry images in certain views, lacking a clear 
structure. Our algorithm successfully preserves morphological consistency across different views and 
accurately captures details, such as the red stripes on the gift box and the clothing details of Mario, 
demonstrating its superior performance.

\begin{figure*}
    \centering
    \includegraphics[width=0.95\textwidth]{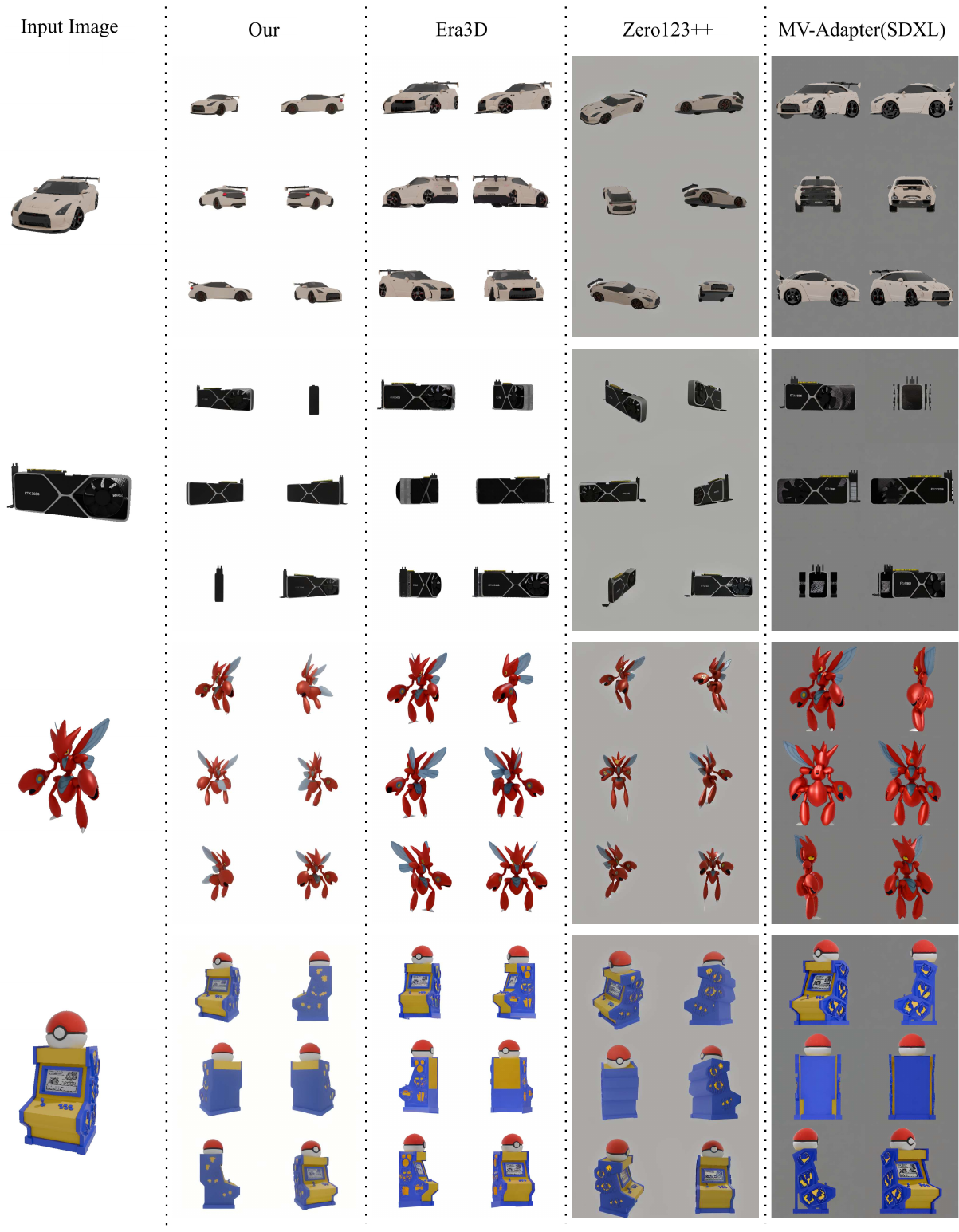}  
    \vspace{-50pt}
    \caption{Comparison of Generation Results on Out-of-domain Images from the Internet between GeoMVD and Other Methods.}
    \label{fig:cmp_web}
\end{figure*}

In Fig.\ref{fig:cmp_web}, we show the results of GeoMVD and other methods on out-of-domain images collected 
from the internet. The comparison highlights the significant advantage of our algorithm in multi-view image 
generation tasks. Our method maintains consistency across views, generating clear, detail-rich images, while 
Era3D and Zero123++ show distortions and blurring in certain views, particularly the back and side views of 
objects. MV-Adapter (SDXL) also suffers from blurriness in detail presentation and fails to effectively represent 
complex structures, especially in the generation of graphics cards and arcade machines. In contrast, our method 
accurately restores the geometric shape and details of objects, displaying higher clarity and consistency. 
Additionally, our approach preserves continuity in shape and detail across multiple views, whereas other methods 
exhibit unnatural deformations and distortions, especially with complex objects such as toy machines and graphics 
cards. Overall, our method shows clear advantages in texture, shape, and viewpoint consistency, producing 
higher-quality multi-view images.

\subsubsection{Quantitative Comparison}

\begin{table}[ht]
    \centering
    \caption{Quantitative Comparison of Image-to-Multi-view Generation on the GSO Dataset between GeoMVD and Other Methods}
    \label{tab:tb1}
    \begin{tabular}{lccc}
        \toprule
        Model & LPIPS $\downarrow$ & FID $\downarrow$ & IS $\uparrow$ \\
        \midrule
        GeoMVD (our)      & \textbf{0.1709} & \textbf{59.0424} & 13.1779 \\
        Era3D             & 0.2261          & 63.3328          & 14.1319 \\
        MV-Adapter (SD2)  & 0.3520          & 85.4668          & \textbf{15.0358} \\
        MV-Adapter (SDXL) & 0.3452          & 77.6760          & 14.3498 \\
        Zero123++         & 0.2581          & 63.7488          & 13.7243 \\
        \bottomrule
    \end{tabular}
\end{table}

We evaluated different models on the validation set (a subset of the GSO dataset) using the 
LPIPS\cite{zhang2018lpips}, FID\cite{heusel2017FID}, and IS\cite{salimans2016IS} metrics. 
To ensure a fair comparison between the methods, we separately rendered the corresponding 
ground-truth images for each method and adjusted the background color accordingly, due to 
variations in background color and viewpoints across the generated images from different models. 
The results are summarized in Table \ref{tab:tb1}. Experimental results show that our GeoMVD 
demonstrates a clear advantage in both generation quality and viewpoint consistency, producing 
higher-quality multi-view images.

\subsection{Ablation Experiment}

To validate the effectiveness of the proposed architecture, we conducted a series 
of ablation experiments to analyze the impact of each key component on the final 
results. By progressively removing the three input features and related components 
upon which the model depends, we can evaluate the contribution of each module to the 
model's overall performance. The experimental design is outlined as follows:

\paragraph{Removal of Cross-modal Conditional Features:} 
To assess the impact of cross-modal features, we set the input CLIP image to a blank 
image and the text prompt to an empty string. This experiment aims to evaluate how 
the absence of cross-modal features affects the consistency and detail preservation 
in the generated outputs.

\paragraph{Removal of Image Conditional Features:} 
In this experiment, we remove the image conditional features, meaning the image no 
longer contributes to the decoupled geometry-enhanced attention calculation. The goal 
is to assess the role of image conditional features in the appearance and detail of 
the generated image, as well as their impact on the denoising process.

\paragraph{Multi-view Geometric Conditional Features:} 
We also conducted a series of ablation experiments on multi-view geometric conditional 
features, which are divided into the following scenarios:

\begin{enumerate}
    \item \textbf{Removal of Geometric Attention Branch:} 
    In this experiment, we completely remove the geometric attention branch, meaning 
    that geometric attention is no longer utilized during the denoising process. 
    The purpose of this experiment is to verify whether the geometric attention 
    branch is crucial for the quality and consistency of the generated images.

    \item \textbf{Removal of Geometric Feature Mask:} 
    In this experiment, we remove the cosine similarity-based dynamic intensity adjustment 
    mechanism, meaning that the intensity of geometric features is no longer adjusted based 
    on the deviation between the target and input views. As a result, the geometric feature 
    intensity for all views remains constant, regardless of their deviation from the main 
    viewpoint. This experiment aims to evaluate the impact of missing geometric feature intensity 
    modulation on the geometric consistency and accuracy of the generated images, as well as the 
    interference of erroneous information on the final results.

    \item \textbf{Removal of Dynamic Geometric Attention Strength Scheduling:} 
    In this experiment, we fix the geometric attention scaling factor $\lambda_{\text{geo}}$, 
    meaning that the strength of this factor is no longer dynamically adjusted based on the 
    time step during the denoising process. Unlike the normal dynamic decay process, the 
    geometric attention scaling factor will remain constant in this experiment. The purpose 
    of this experiment is to assess the impact of the absence of dynamic decay in the geometric 
    attention scaling factor on the final results, particularly whether a high fixed geometric 
    attention factor interferes with image details and negatively affects the quality and 
    consistency of the generated image.
\end{enumerate}

\subsubsection{Quantitative Comparison of Ablation Experiments}

Table\ref{tab:tb2} presents the quantitative comparison results of different ablation experiments. 
The experimental results clearly show that geometric attention, geometric feature masks, 
and dynamic geometric attention strength scheduling play crucial roles in the quality 
of the generated images. Removing these modules leads to significant degradation in 
consistency, detail, and diversity of the generated images. In particular, the 
interaction between image conditional features and geometric conditional features, 
the modulation of geometric feature intensity, and the dynamic adjustment of the 
attention mechanism are all essential for the denoising process. These experiments 
validate the effectiveness of the proposed architecture and provide a basis for 
further optimization.
Qualitative Comparison of Ablation Experiments can be found in the appendix \ref{sec:appendix_section}.


\begin{table}[tbh]
    \centering
    \caption{Quantitative Comparison in Ablation Experiments}
    \label{tab:tb2}
    \small  
    \begin{tabularx}{\linewidth}{Xccc}  
        \toprule
        Ablation Methods                                      & LPIPS $\downarrow$ & FID $\downarrow$ & IS $\uparrow$ \\
        \midrule
        GeoMVD (Our)                                          & \textbf{0.1709}    & \textbf{59.0424} & 13.1779          \\
        w/o Cross-modal Conditional Features                  & 0.1934             & 67.8664          & \textbf{13.9451} \\
        w/o Image Conditional Features                        & 0.3075             & 131.8195         &  9.1374          \\
        w/o Geometric Attention Branch                        & 0.1795             & 61.5617          & 12.7903          \\
        w/o Geometric Feature Mask                            & 0.1760             & 61.1665          & 12.8486          \\
        w/o Dynamic Geometric Attention Strength Scheduling   & 0.2624             & 83.8276          & 13.2391          \\
        \bottomrule
    \end{tabularx}
\end{table}

\subsubsection{Analysis and Discussion}

The following is an analysis and discussion of the results from each experimental 
setup:

\paragraph{Removal of Cross-modal Conditional Features:} 
Cross-modal conditional features, which combine image and text, provide essential 
cross-modal guidance within the model. After removing these features, the consistency 
of the generated images across different views significantly deteriorates, resulting in 
distortions and unnatural effects. This emphasizes the critical role of cross-modal 
features in maintaining the consistency of generated images. The combination of image 
and text not only provides richer semantic information but also reduces distortions 
during multi-view generation, ensuring the consistency of the object’s shape. Thus, 
cross-modal features are indispensable for the naturalness and consistency of 
generated images across different views.

\paragraph{Removal of Image Conditional Features:} 
When the image conditional features are removed, the model is unable to accurately 
understand or restore the object’s shape and details, leading to a complete degradation 
of the generated outputs. The object loses its original structure, shape, detail, and 
consistency. This result underscores the crucial role of image conditional features in 
preserving the appearance, detail, and geometric consistency of the object. Without 
image guidance, the overall quality of the generated image is significantly reduced, 
leading to a loss of realism and detail. This highlights the importance of image 
conditional features for the quality of the generated results.

\paragraph{Geometric Features and Condition Analysis:}  
The role of geometric features in multi-view generation models is crucial and manifests in several key aspects: 

\begin{itemize}
    \item \textbf{Shape Consistency:} 
    Geometric features ensure that the object maintains consistent shape and structure across 
    different views, preventing distortions or inconsistencies in shape between views.
    \item \textbf{Structural Coherence:} 
    Geometric features help maintain correct occlusion relationships and relative positioning 
    across views, ensuring that the object appears natural from all perspectives.
    \item \textbf{Detail Restoration:} 
    During the denoising process, geometric features provide high-frequency information such 
    as edges and contours, aiding in the restoration of object details and enhancing clarity 
    and precision in the generated image.
\end{itemize}

The effectiveness of geometric conditions depends on their strength. When the geometric condition is 
too weak, overly strong, or improperly controlled, it significantly impacts the generation quality of 
multi-view images. Specifically:

\begin{enumerate}
    \item \textbf{Weak Geometric Condition Strength (Removal of Geometric Attention Branch):} 
    When the geometric condition is too weak, the consistency of shape and structure across 
    different views deteriorates significantly. After removing the geometric attention branch, 
    the object’s structure and shape fail to remain consistent across multiple views, especially 
    in the generation of complex objects, where the shape appears blurry and fragmented. The lack 
    of necessary geometric constraints leads to a loss of structure and spatial coherence in the 
    generated images.

    \item \textbf{Improper Geometric Condition Strength Control (Removal of Geometric 
    Feature Mask):} The geometric feature mask adjusts the intensity of geometric 
    information across different views, ensuring shape consistency. After removing 
    the geometric feature mask, the geometric structure of the object fails to align 
    correctly across different views, particularly in the generation of complex objects. 
    Occlusion relationships and shape deviations emerge, causing geometric 
    inconsistencies in the generated images. Without the geometric feature mask, 
    the object’s structure becomes incoherent, which negatively affects image quality.

    \item \textbf{Excessive Geometric Condition Strength (Removal of Dynamic Geometric 
    Attention Strength Scheduling):} 
    The geometric feature mask adjusts the intensity of geometric information across 
    different views, ensuring shape consistency. After removing the geometric feature mask, 
    the geometric structure of the object fails to align correctly across views, especially 
    for complex objects. Occlusion relationships and shape deviations emerge, causing geometric 
    inconsistencies in the generated images. Without the geometric feature mask, the object’s 
    structure becomes incoherent, which negatively impacts image quality.
    
\end{enumerate}

In conclusion, geometric conditions play a crucial role in generating multi-view images by 
ensuring shape consistency, structural coherence, and detail restoration. When geometric 
conditions are too strong, the generated images suffer from excessive constraints, leading 
to distortions and unnatural details. When geometric conditions are too weak, the lack of 
consistency in object shape causes the images to lose coherence and realism across views. 
Improper control of geometric information results in geometric inconsistencies, where the 
generated images fail to align across views. Proper adjustment of geometric condition strength 
is key to generating high-quality images. By applying appropriate geometric constraints, the 
model ensures shape consistency while restoring details and maintaining the naturalness and 
accuracy of the images.

\section{Conclusion}
\label{sec:conclusion}

This paper proposes a Geometry-guided Multi-view Diffusion Model (GeoMVD) designed to address the challenge 
of generating consistent multi-view images from a single input image. We introduce several key components, 
including a multi-view geometric detail extraction module, a conditional feature extraction process, a decoupled 
geometry-enhanced attention mechanism, and a geometric information intensity modulation mechanism. These 
components significantly enhance the quality, consistency, and detail fidelity of the generated images. 
Experimental results on multiple datasets demonstrate that GeoMVD outperforms existing methods, particularly 
in terms of detail preservation and cross-view consistency.

However, there are still some limitations in this study. First, while the multi-view geometric detail 
extraction module effectively resolves the issue of geometric consistency, the process remains relatively 
complex and computationally expensive. Second, the model's adjustment of geometric features may not be 
sufficiently fine-tuned for complex scenes, limiting the naturalness and detailed representation of the 
generated images.

Future work will focus on several areas: (1) improving the efficiency and accuracy of geometric information 
extraction to reduce computational costs, (2) enhancing the model's ability to generalize and adapt to a 
broader range of scenes and object types, and (3) expanding the integration of multi-modal information to 
mprove the model’s cross-domain applicability. Through these efforts, we aim to further advance the practical 
application of multi-view generation models, especially in fields such as virtual reality, augmented reality, 
and autonomous driving.

{\small
\bibliographystyle{ieeenat_fullname}
\bibliography{11_references}
}

\ifarxiv \clearpage \appendix \section{Qualitative Comparison of Ablation Experiments}
\label{sec:appendix_section}

\paragraph{Removal of Cross-modal Conditional Features:} 
After removing the cross-modal conditional features, the quality of the generated multi-view images 
significantly decreases. As shown in Fig.\ref{fig:wo_cross}, the morphological consistency of the 
object across different views is compromised, with some views exhibiting shape distortions and improper 
structural alignment. For instance, the details of the sofa and chair, particularly in terms of 
texture and fabric representation, are lost, severely affecting the realism of the generated images. 
Additionally, removing the cross-modal conditional features disrupts the consistency of color and 
texture, making the generated objects appear visually unnatural. These results highlight the crucial 
role of cross-modal conditional features in maintaining consistency and detail fidelity in multi-view 
image generation.

\paragraph{Removal of Image Conditional Features:} 
After removing the image conditional features, the generated multi-view images lose their original 
consistency and detail restoration capabilities. As shown in Fig.\ref{fig:wo_img}, in the absence of 
image conditional features, the generated objects exhibit significant distortion and inconsistency in 
both shape and texture across multiple views. The details of the objects, such as edges, contours, 
and surface structure, fail to maintain coherence across different viewpoints, resulting in blurry 
images that lack realism. The accuracy of the overall shape and the fidelity of details notably decrease, 
particularly in the reconstruction of complex structures and surface textures. These results demonstrate 
that image conditional features are essential for ensuring consistency and detail recovery in generated 
images, and their absence leads to a substantial decline in the quality of the generated results.

\paragraph{Removal of Geometric Attention Branch:} 
As shown in Fig.\ref{fig:wo_geo_atten}, after removing the geometric attention branch, the quality of 
the generated multi-view images significantly declines in terms of shape consistency and detail restoration. 
The geometric structure of the object fails to remain consistent across different views, resulting in 
distorted shapes and unnatural variations. Occlusion relationships and the relative positions of the 
objects are also not properly coordinated across views, leading to a lack of spatial coherence in the 
generated results. Moreover, the performance of finer details is unstable, with a weakened ability to 
restore object edges, textures, and small structures, resulting in blurred and distorted outputs. The 
absence of the geometric attention branch weakens the geometric constraints on the image, causing 
inconsistencies in shape and details across multiple views. Therefore, the geometric attention branch 
plays a critical role in ensuring the consistency, detail fidelity, and structural coherence of multi-view 
images, and its removal significantly impacts the quality and naturalness of the generated results.

\paragraph{Removal of Geometric Feature Mask:} 
After removing the geometric feature mask, the quality of the generated multi-view images significantly 
deteriorates. As shown in Fig.\ref{fig:wo_geo_mask}, the consistency of the object's shape and details 
is compromised, particularly in maintaining the geometric structure and occlusion relationships. The 
shapes of the generated backpack and boots are inaccurate, especially on the back and bottom of the 
backpack, where the shape becomes distorted. The geometric feature mask ensures consistency in shape 
and occlusion relationships across different views by adjusting the intensity of the geometric information. 
Without this mask, the structure and shape of the objects fail to align properly across views, resulting in 
geometric inconsistencies in the generated outputs. Furthermore, the generated basket and boots lose their 
structural integrity across multiple views, with severe detail loss, especially in surface and fine details. 
Without the support of the geometric feature mask, the structure of the objects becomes blurred. Overall, 
the geometric feature mask provides stable geometric constraints during the generation process, ensuring 
the 3D feel and spatial coherence of the objects. Its removal leads to a lack of consistency in the shape 
and structure of the generated images.

\paragraph{Removal of Dynamic Geometric Attention Strength Scheduling:} 
As shown in Fig.\ref{fig:wo_dynamic_geo}, after removing the dynamic geometric attention strength scheduling, 
the generated multi-view images exhibit noticeable issues in shape consistency and detail restoration. The 
geometric structure and surface details of the object fail to remain consistent across different views, 
resulting in distortions in shape and structure between multiple views. Notably, the handling of surface 
details and textures in the generated images shows unnatural gloss and blurriness, particularly in the 
finer details. This indicates that dynamic geometric attention strength scheduling is crucial for ensuring 
geometric consistency, detail fidelity, and structural coherence across different views. Its absence leads 
to deviations in object shape and loss of detail, negatively impacting the quality and naturalness of the 
generated images.

\begin{figure*}[b]
    \centering
    \includegraphics[width=0.95\textwidth]{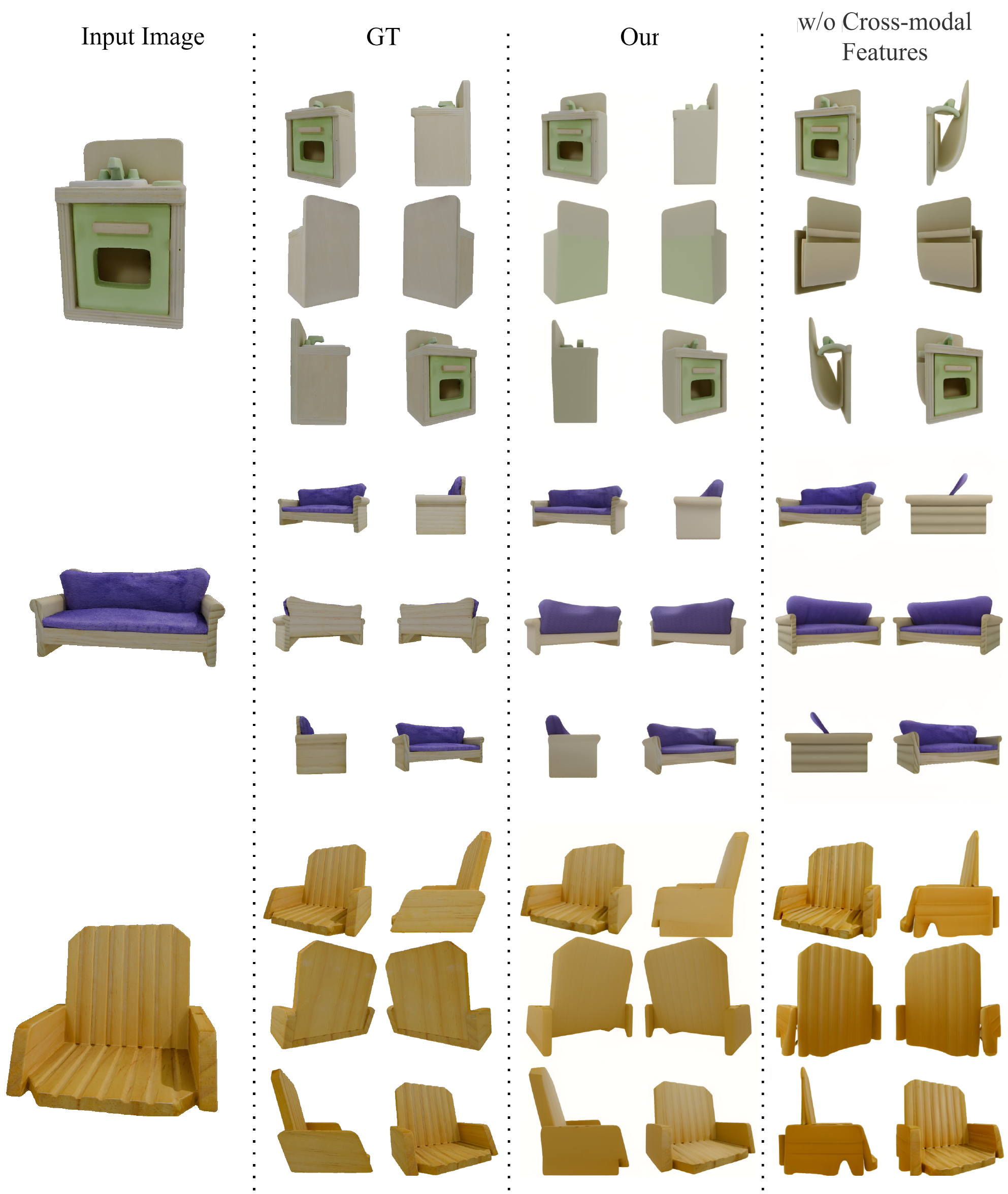}  
    \caption{Removal of Cross-modal Conditional Features.}
    \label{fig:wo_cross}
\end{figure*}

\begin{figure*}
    \centering
    \includegraphics[width=0.95\textwidth]{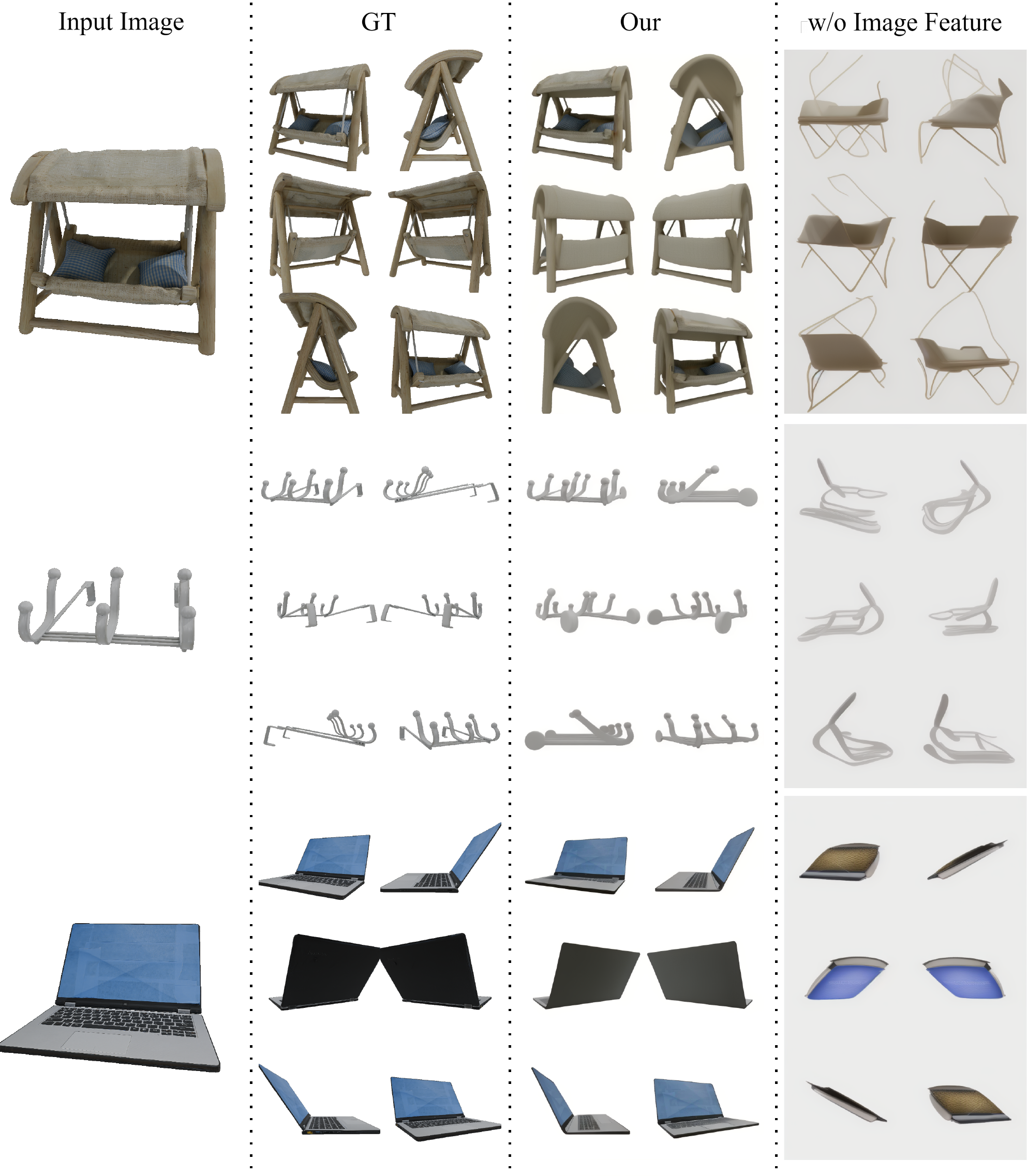}  
    \caption{Removal of Image Conditional Features.}
    \label{fig:wo_img}
\end{figure*}

\begin{figure*}
    \centering
    \includegraphics[width=0.95\textwidth]{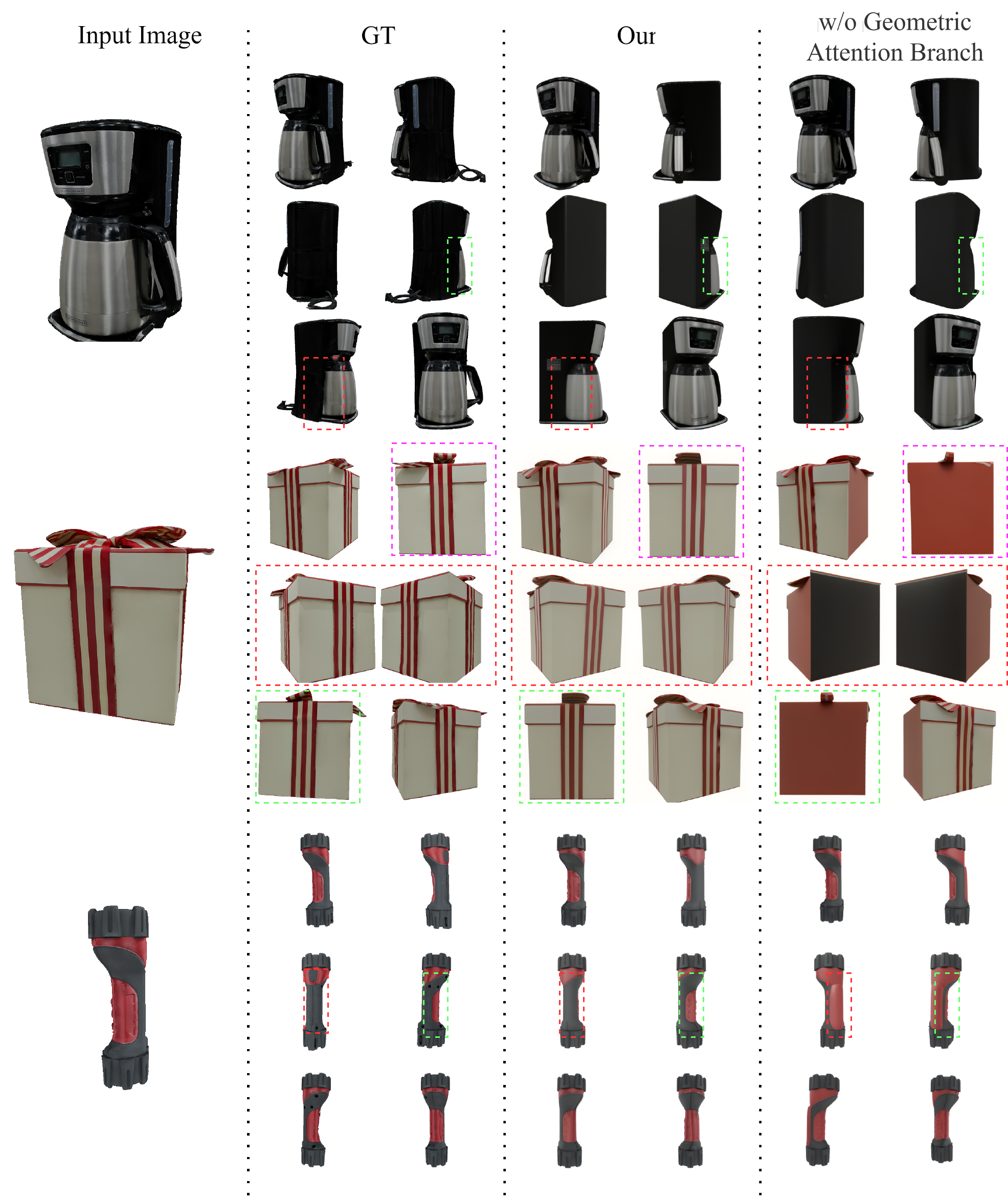}  
    \caption{Removal of Geometric Attention Branch}
    \label{fig:wo_geo_atten}
\end{figure*}

\begin{figure*}
    \centering
    \includegraphics[width=0.95\textwidth]{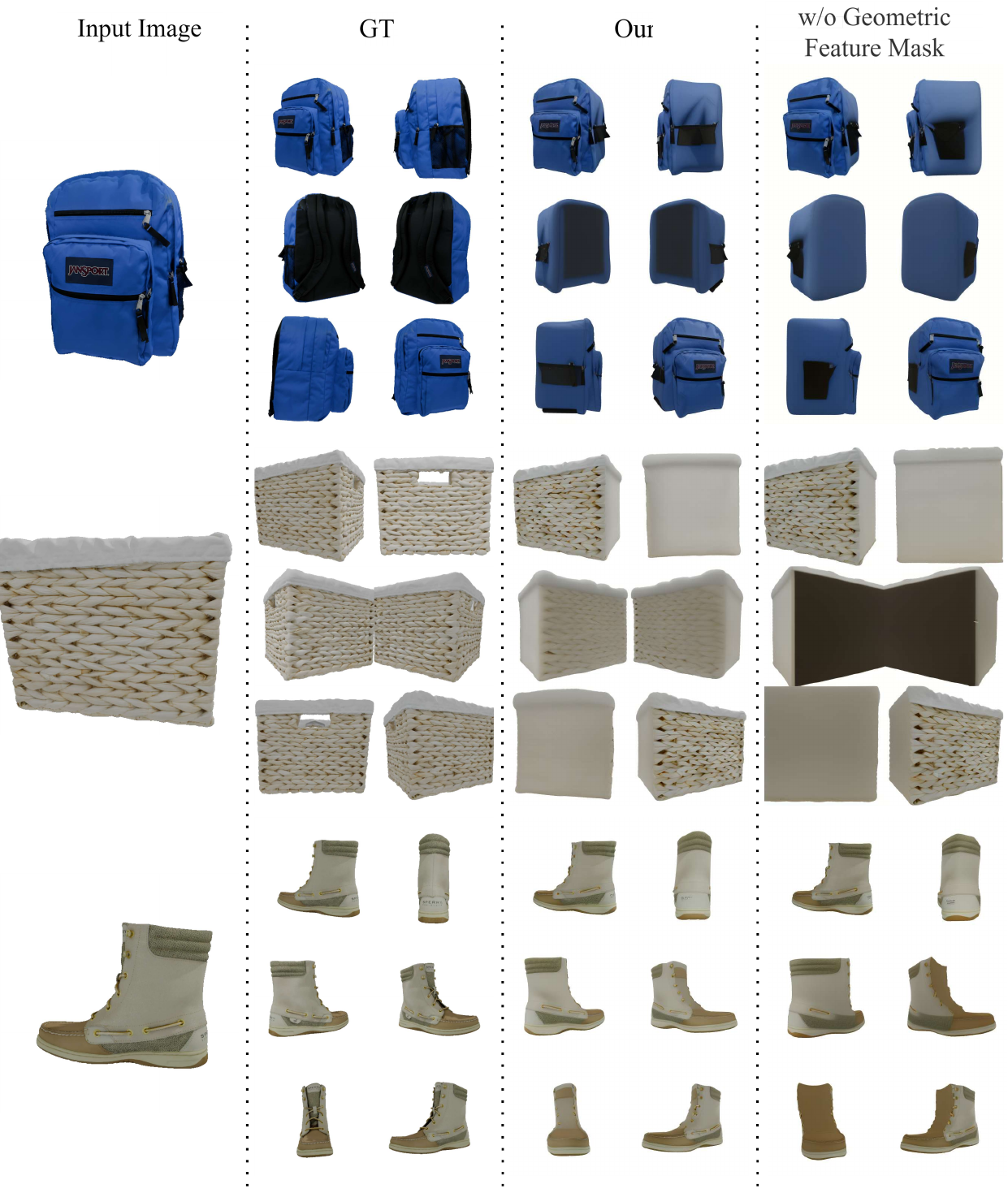}  
    \vspace{-50pt}
    \caption{Removal of Geometric Feature Mask}
    \label{fig:wo_geo_mask}
\end{figure*}

\begin{figure*}
    \centering
    \includegraphics[width=0.95\textwidth]{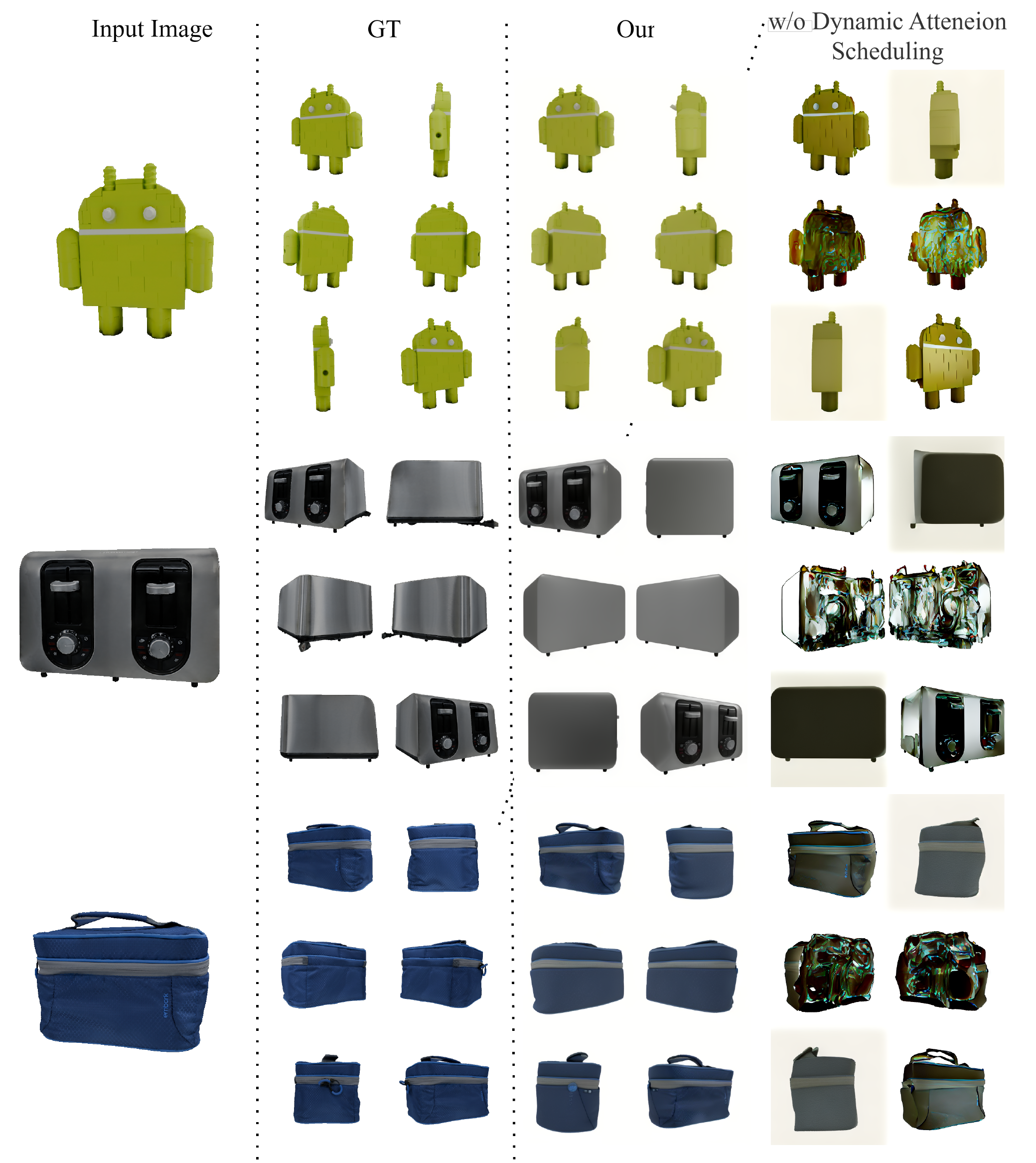}  
    \caption{Removal of Dynamic Geometric Attention Strength Scheduling}
    \label{fig:wo_dynamic_geo}
\end{figure*}

 \fi

\end{document}